\documentclass[12pt]{article}

\usepackage{scicite}

\usepackage{times}
\usepackage{graphicx}
\usepackage{subfig}
\usepackage{xspace}

\usepackage{hyperref}

\newcount\Comments  
\usepackage{color}
\definecolor{darkgreen}{rgb}{0,0.5,0}
\definecolor{purple}{rgb}{1,0,1}
\definecolor{teal}{rgb}{0,0.4627,0.5804}
\newcommand{\kibitz}[2]{\ifnum\Comments=1\textcolor{#1}{#2}\fi}

\newenvironment{sciabstract}{%
\begin{quote} \bf}
{\end{quote}}

\title{Enabling Mixed Autonomy Traffic Control}

\author{Matthew Nice$^{1\ast}$, 
Matt Bunting$^{1}$, 
Alex Richardson$^{1}$, \\
Gergely Zachàr$^{1}$, 
Jonathan W.~Lee$^{2}$, 
Alexandre Bayen$^{2}$, \\
Maria Laura {Delle Monache}$^{2}$, 
Benjamin Seibold$^{3}$, 
Benedetto Piccoli$^{4}$, \\
Jonathan Sprinkle$^{1}$, and 
Dan Work$^{1}$ \\
\normalsize{$^{1}$ Vanderbilt University, Nashville, TN, USA}\\
\normalsize{$^{2}$ University of California , Berkeley, Berkeley, CA}\\
\normalsize{$^{3}$ Temple University, Philadelphia, PA}\\
\normalsize{$^{4}$ Rutgers University at Camden, Camden, NJ}
\\
\normalsize{$^\ast$Corresponding author. E-mail:  matthew.nice@vanderbilt.edu.}
}

\date{}

\usepackage{palatino}

\begin{document}


\baselineskip24pt


\maketitle

\begin{sciabstract}
  We demonstrate a new capability of automated vehicles: mixed autonomy traffic control. With this new capability, automated vehicles can shape the traffic flows composed of other non-automated vehicles, which has the promise to improve safety, efficiency, and energy outcomes in transportation systems at a societal scale. Investigating mixed autonomy mobile traffic control must be done in situ given that the complex dynamics of other drivers and their response to a team of automated vehicles cannot be effectively modeled. This capability has been blocked because there is no existing scalable and affordable platform for experimental control. This paper introduces an extensible open-source hardware and software platform, enabling a team of 100 vehicles to execute several different vehicular control algorithms as a collaborative fleet, composed of three different makes and models, which drove 22752 miles in a combined 1022 hours, over 5 days in Nashville, TN in November 2022.

\end{sciabstract}


\section*{Introduction}

This article introduces a new capability for connected automated vehicles (CAVs): mixed autonomy traffic control. Mixed autonomy means that a proportion of the vehicles are automated, and the remainder are not. Even when automated vehicles are a small fraction of the flow, mixed autonomy traffic control has the potential for significant societal-scale benefits~\cite{talebpour2016influence,delle2019autonomous,forster2014cooperative,stern2018dissipation}, e.g. reducing energy in stop-and-go driving by up to 40\%\cite{stern2018dissipation}.

In this work we deploy a fleet of CAVs 'in the wild', i.e. in a mixed autonomy setting on the freeway. The CAV fleet exchanges information to work collaboratively on a shared goal. Existing objectives for CAV fleets include platooning~\cite{chang1991experimentation,al2010experimental} to reduce aerodynamic drag for the vehicles within the platoon, and operating as efficient automated ride sharing networks~\cite{zhao2020enhanced}. This article takes CAV fleets in a new direction, by deploying a CAV fleet with the objective to reduce phantom traffic jams~\cite{gazis1959car,chandler1958traffic} by controlling traffic flow in the mixed autonomy setting. 

Researchers want investigate the promise of robotics in mixed autonomy traffic settings, but there is a fundamental conflict. These technologies need to be investigated at scale \textit{in situ} because there is no holistic replacement for the real physical traffic environment. Simultaneously, potential technologies have not yet been investigated \textit{in situ} because of the issues of scale necessarily introduced to create and measure their impact; there is a gap in the tools and technologies to deploy CAVs at scale. To address this gap, we develop new scalable hardware and software that endows commodity vehicles with the ability to sense the traffic environment, perform vehicle control in a coordinated manner, and adapt to a complex field environment.


There are relatively few large-scale robotic passenger vehicle deployments in the field, because of the inherent difficulty of deploying experimental technologies at scale. The most visible examples are the highly automated vehicle fleets that use expensive suites of sensors and onboard computation in pursuit of SAE level 4+ driving. Our work differs in that the goal of our fleet is to change the emergent properties of traffic. We demonstrate deployment can be achieved at scale with an approach using low cost computation and stock sensors on today's commodity vehicles. The capability of mixed autonomy traffic control has been anticipated for decades~\cite{nice2021can,vinitsky2018lagrangian,work2009lagrangian,herrera2010incorporation,kato2002vehicle,shladover1991automated,lichtle2022deploying}. Our low cost approach addresses the gap in tools and technologies, and enables the deployment of mixed autonomy traffic control with exchangeable candidate experimental control.

\paragraph*{Contributions}

    \begin{enumerate}
            \item We deploy the first large-scale team of connected automated vehicles (CAVs) for mixed autonomy traffic control.
        \item We introduce a hardware and software platform to enable experimental autonomy and connectivity in  commercially available SAE level 1 and level 2 automated vehicles. The platform supports experimental automated vehicle control, live mobile sensing with connected vehicles, high fidelity data collection, and scalability.
        \item This platform enables an agile develop/deploy cycle for at scale cyber-physical systems research, empowering its field deployment. 

    \end{enumerate}
\section*{Results}

\begin{figure}
    \centering
    \href{https://youtu.be/slH9nimpaY8}{\includegraphics[width=1.0\linewidth]{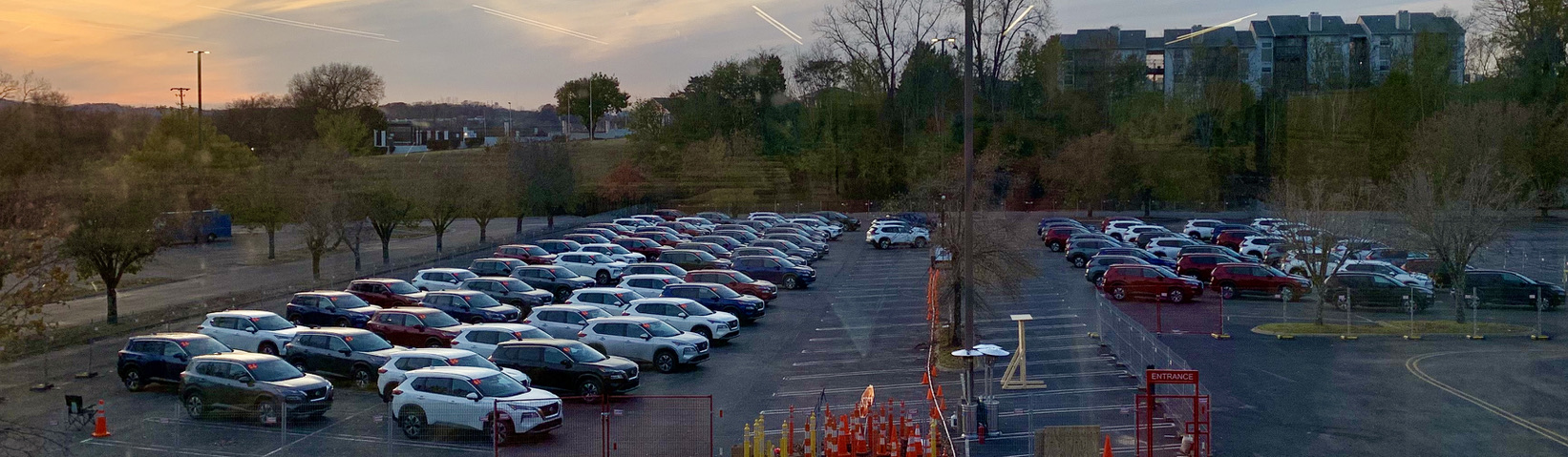}}
    \caption*{Movie 1: Deploying mixed autonomy traffic control with a fleet of 100 connected automated vehicles. Available at \href{https://youtu.be/slH9nimpaY8}{https://youtu.be/slH9nimpaY8}}
\end{figure}

This work introduces the tools and technologies needed to transform a fleet of commodity vehicles into a team of connected automated vehicles. With these tools in place, we are able to deploy mixed autonomy traffic control in the wild with our CAV fleet. The demonstration of the CAV fleet circulates on a 5-mile loop of Interstate 24 (I-24) near Nashville, TN, a multi-lane freeway with heavy congestion. Each of the SAE Level 1-2 vehicles is augmented with an experimental control system that driver operators activate on the freeway in the same manner as existing cruise control systems. Leveraging their connectivity, the vehicles run collaborative control algorithms with the objective to smooth traffic waves in morning congestion. The results evaluate the effectiveness of our platform to create a scaled CAV fleet; we do not evaluate the effectiveness of specific exchangeable algorithms. First, we consider the fleet as a coordinated CAV team actuating on the congested traffic flow. Then, we analyze the deployment for control vehicle density and approximate penetration rate. Last, we introduce the novel tools and technologies which enable a single vehicle to become cohesive with others at scale. 

\begin{figure}
        \centering
        \includegraphics[width=1.0\linewidth]{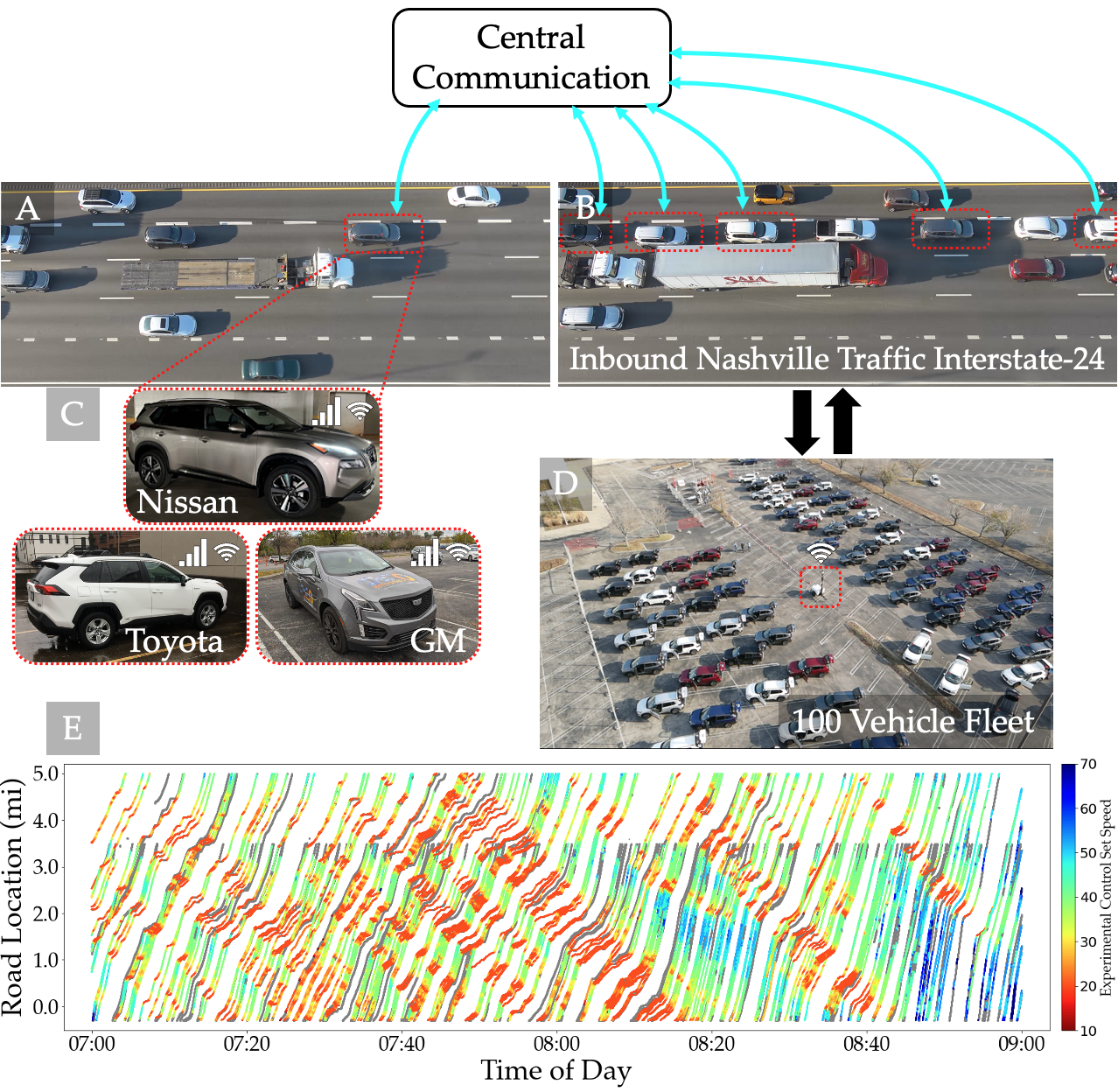}
        \caption{\textbf{Collaborative Automated Fleet as Mobile Agents on Traffic Flow.} The ephemeral fleet of 100 experimental vehicles (\textbf{D}) was deployed in varying traffic densities (\textbf{A-B}). Drivers circulated in congested morning traffic, and when they needed a break the vehicle had a fresh driver swapped in at our operational HQ (D). The heterogeneous fleet had 3 constituent vehicle platforms from 3 different major manufacturers (\textbf{C}). (\textbf{E}) shows the series of inbound Nashville trajectories from our large-scale experimental fleet over a 5 mile stretch of Interstate-24 in morning congestion. When experimental control is engaged, the commanded vehicle speed is shown according to the color bar; otherwise, the data point is colored gray. }
        \label{fig:fig1}
    \end{figure}
   \paragraph{A Fleet of 100 Mobile Agents on the Traffic Flow.}

   We deploy the fleet of CAVs on I-24 during heavy morning congestion from November 14-18, 2022. The CAV fleet runs experimental control algorithms distributed densely on a five mile stretch of roadway as shown in Figure~\ref{fig:fig1}. By extending commodity vehicles into a collaborative fleet of 100 connected automated vehicles, we are able to test in the wild for the first time mixed autonomy traffic control at scale. Our fleet drove a cumulative 22752 recorded miles (36616 kilometers) over a combined 1022 hours. In the field, tens of versions of software are deployed, with 6 deployed on 50+ vehicles and 19 deployed on 10+ vehicles. The deployed controllers are designed for distributed traffic wave smoothing control. 

   Our vehicle fleet faced varied traffic densities (Figure~\ref{fig:fig1}~A-B) in recurring congestion while running experimental mixed autonomy traffic control algorithms. The 100 vehicle fleet consists of 3 vehicle models from 3 original equipment manufacturers (OEMs) (Figure~\ref{fig:fig1}~C). The variation in OEMs required cross-platform support from our hardware and software tools. Vehicle-specific hardware was abstracted away with our lower-level software, providing a uniform interface for upper-level software modules~\cite{nice2023middleware,bunting2021libpanda}, including modular ROS integration for control developers and live tracking~\cite{richardson2023analysis} for field test administrators. Vehicles circulated in traffic flow (Figure~\ref{fig:fig1}~A-B) until driver operators elected to rest at our operational headquarters (HQ) (Figure~\ref{fig:fig1}~D), providing an opportunity for fresh operator to continue circulating the vehicle. The HQ parking lot WiFi access point (Figure~\ref{fig:fig1}~D, in red), from a spliced extension of the building's fiber optic network, enabled high volume data transfer each time the vehicles are powered off. This network connection allowed for immediate data backup and data access to analyze control performance, which informed new software versions installed over-the-air on the next startup of the vehicles. Strym~\cite{bhadani2022strym} and ROS~\cite{quigley2009ros} were key data analysis tools in ad-hoc and urgent analyses of in-vehicle network data informing field decisions. The CAV fleet was distributed throughout the inbound Nashville morning congestion in a 5 mile stretch of freeway. Figure~\ref{fig:fig1}~E shows 323 control vehicle trajectories, and where experimental control was actively commanding the vehicle, from 7AM-9AM on the test day. The varied colors indicate the commanded velocity, and gray indicates where the system was not active. Notice that for driver operators which turn around at MM 3.5 there is a consistent gray tip of their trajectory; this is a result of electing to disengage the experimental control system as the operator prepares to exit the highway. As they drive around, each CAV in the team shares information with our central server, which allows central control whitelisting, live tracking, control heartbeats, local sensor message passing, control recommendations and more; see the materials and methods section for more details. 

\paragraph{A Deployment of Mixed Autonomy Traffic Control Examined.}

\begin{figure}
        \centering
        \includegraphics[width=\linewidth]{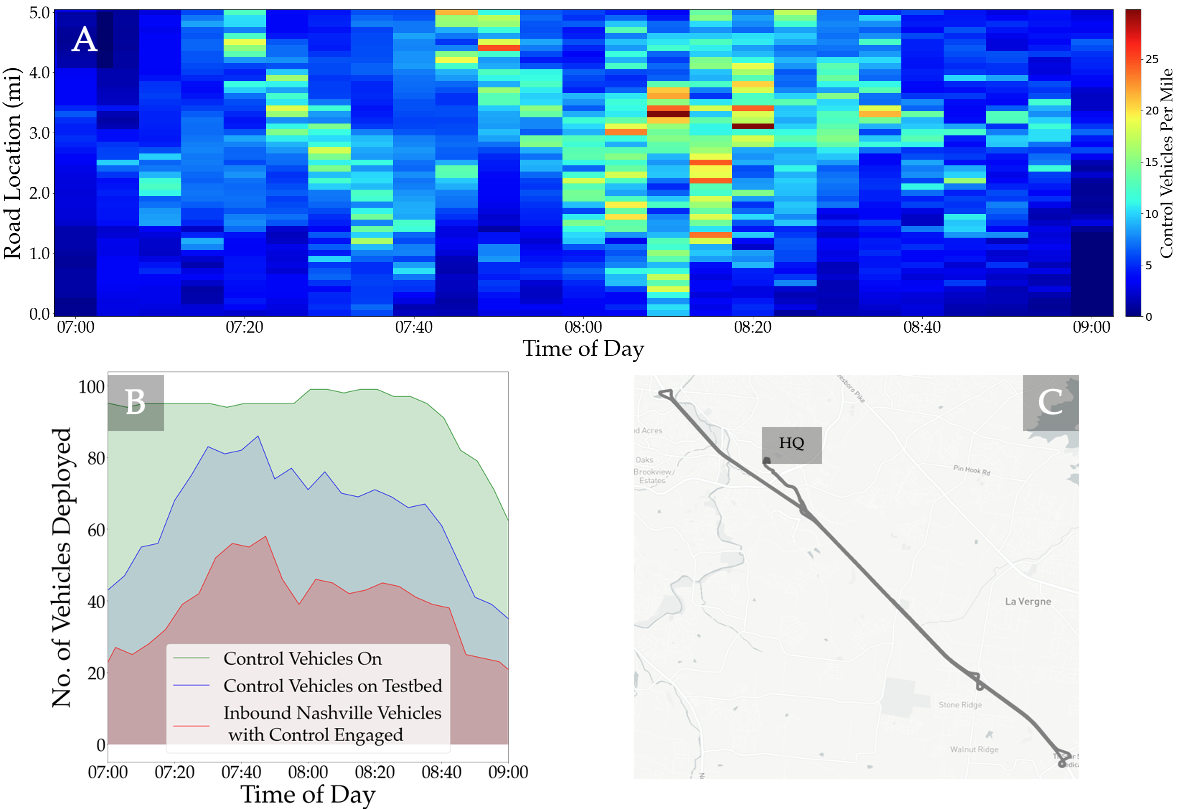}
        \caption{\textbf{Control Fleet Deployment Examined.} The control vehicle density (\textbf{A}) is created by calculating the density of westbound (inbound Nashville) vehicles in 528 ft (0.1 mile) and 5 minute windows. Counts of control vehicles (\textbf{B}) has three categories compared. A vehicle is considered "on" once the data starts recording. A vehicle is "on the testbed" if it is along the experimental corridor (not including turning around at exit/entrances to I-24). Inbound Nashville vehicles with control engaged are counted by the number of inbound (i.e. westbound I-24) vehicles within a 5 minute window that have control engaged. The partially overlapping looping routes driven by control vehicles, to support penetration rate and avoid overcrowding on/off ramps, are shown in (\textbf{C}).}
        \label{fig:fig2}
    \end{figure}

    To examine the mixed autonomy traffic control deployment, we examine the control vehicle density, control vehicle penetration rate in vehicle flow, vehicle volume, and looping I-24 driving routes. Vehicle density is a measure of the number of vehicles per roadway length, vehicle flow is a measure of the number of vehicles moving per unit of time, and the penetration rate is the proportion of vehicles of interest. Figure~\ref{fig:fig2}~A shows the density of control vehicles per mile throughout the 5 mile experimental corridor and morning of congestion. Density is calculated per 5 minute and 0.1 mile (528 ft) window. Vehicle flow during heavy congestion on I-24 ranges from 6000-8000 vehicles per hour across all lanes, which translates to a 2.0\%-2.7\% average penetration rate for our control vehicles from 7AM-9AM on Friday 11/18/2023 across all lanes of traffic. If we consider just the three lanes of four total on the roadway in which control vehicles were operating, the average penetration rises to a range of 2.7\%-3.6\%. This penetration rate could be sufficiently high for a distributed control system to effect the flow of traffic for all vehicles in the 5 mile experimental corridor. Previous work on a closed course ring road used 1 automated control vehicle and 20 non-automated vehicles (4.8\%), i.e. 1 control vehicle per 260m (853 ft) ring circumference and achieved dramatic traffic smoothing behavior~\cite{stern2018dissipation}; this density indicator to helps inform our understanding of approximate effective densities, but can not be compared directly to effects in large-scale deployment in the wild.

    For reference on the scale of distance on the interstate, at highway speeds (70~mph) the vehicle a driver is following at a standard $\sim$3s time gap is $\sim$300 feet away; that is 17.2 vehicles per mile. The local density of our control vehicles per mile reached a maximum above 26 vehicles per mile.  Standing on the side of Interstate-24 observing this deployment, one would see an inbound Nashville control vehicle from our CAV fleet passing on average every 22 seconds. The precise density-effectiveness trade-offs of CAVs in mixed autonomy traffic control in real settings requires more research outside the scope of this work.


    To understand our CAV deployment further, we can look at the status of all 100 vehicles in Figure~\ref{fig:fig2}B, and consider the maximums they reached. The total number of vehicles concurrently running our software reaches a maximum of 99, the number of vehicles on the experiment corridor concurrently reaches a maximum of 86, and the number of simultaneously active control vehicles in the inbound Nashville morning traffic reaches a maximum of 58, as shown in Figure~\ref{fig:fig2}B. The vehicles in the fleet operate in a loop, shown in Figure~\ref{fig:fig2}C; this tactic supports the control vehicle density, and therefore the effect in traffic, in being out-sized even compared to the large fleet. 

\begin{figure}
        \centering
        \includegraphics[width=0.78\linewidth]{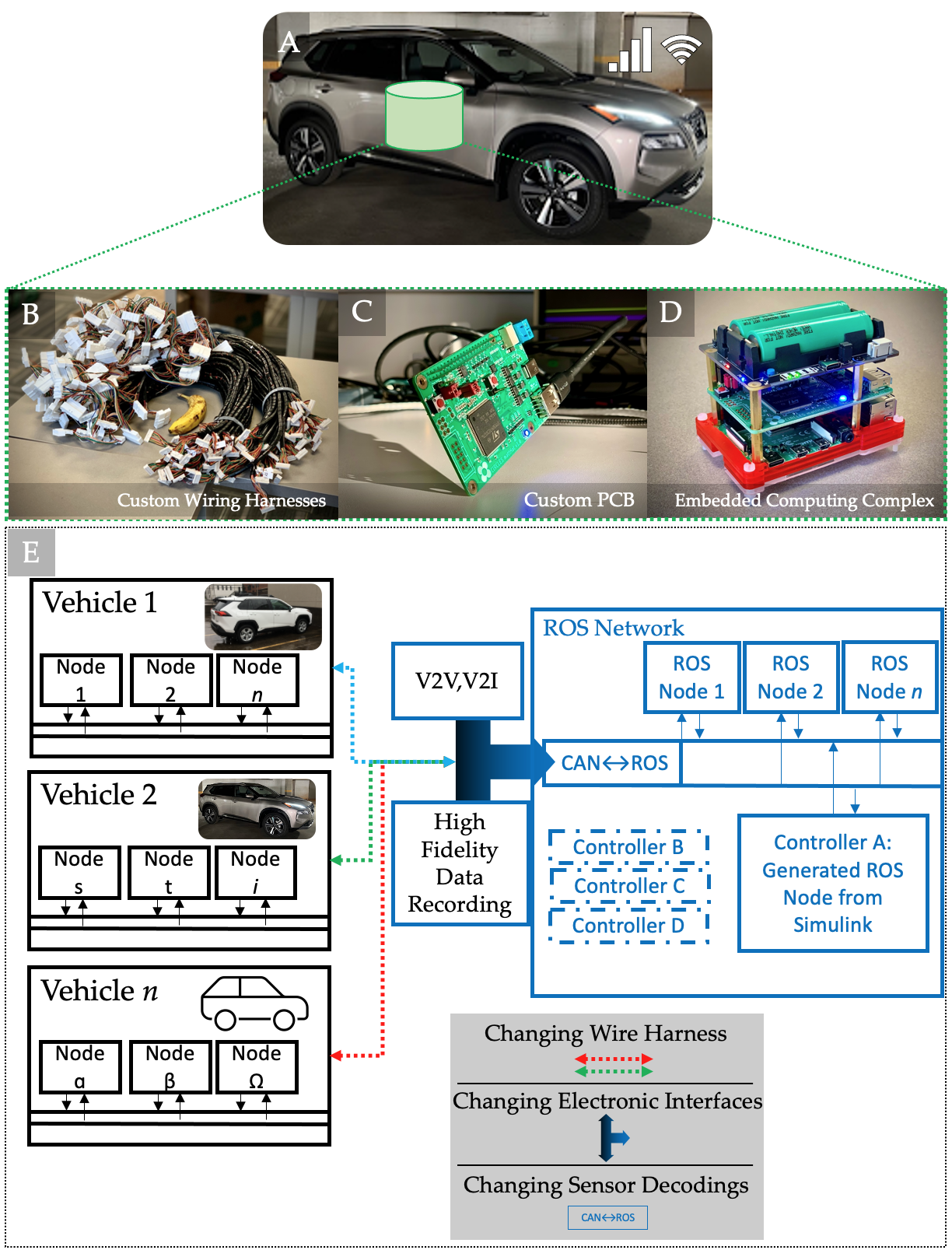}
        \caption{\textbf{Infrastructure of a Scalable Connected Automated Vehicle.} Overview of the hardware and software infrastructure within a vehicle (\textbf{A)}. The key hardware components (\textbf{B-D)} are installed in each vehicle to bridge an extension of the stock vehicle electronic system to experimental control. (\textbf{E}) portrays the 2-way bridge between heterogeneous in-vehicle networks and the ROS framework. This allows the system to leverage existing infrastructure like Simulink ROS node generation for controller development, deployable immediately in the heterogeneous CAV team. Additionally, there is integration with inter-vehicle communication, and data recording, among other apps.} 
        \label{fig:fig3}
\end{figure}

\paragraph{A Scalable Connected Automated Vehicle.}
Creating a scaled CAV fleet starts with creating a modification strategy for new connectivity and control in a single vehicle that is scalable. A series of significant pieces of hardware and software infrastructure are needed in order to extend a commodity vehicle into a connected automated vehicle. With regard to hardware: our custom wiring harness and printed circuit boards (PCB) create an electronic-to-digital bridge between a vehicle's physical sensing and actuation network and our embedded computer; the PCB, embedded computer, and battery uninterrupted power supply constitute the embedded computing complex (Figure~\ref{fig:fig3}~B-D). A custom wiring harness is necessary because unlike OBD-II interfaces, there is not a standardized connector and wiring layout for in-vehicle networks. A custom PCB was needed to overcome price constraints and scarcity in the embedded computer chip market, as well as provide the lean functionality we required. The need for lean principles in implementation cannot be overstated; simultaneously the specifications for a functional field test require a high level of robustness and agility in implementation.

To mediate the information sharing two-ways between the vehicle and experimental control, a suite of novel software tools were created or extended. Notably we extend earlier work: libpanda\cite{bunting2021libpanda}, a broad-scope package which houses data recording and vehicle control interfaces, and CAN-to-ROS\cite{elmadani2021can,nice2023middleware} a model-based code generative package providing a dynamic bridge between heterogeneous vehicle networks and the ROS framework for real-time sensor data and control. Other software components were created or extended to support and leverage information sharing from vehicle to infrastructure (critical for scientific understanding and execution of field testing) and vehicle to vehicle (enabling fleet collaboration, facilitated by central server proxy). These hardware and software pieces are discussed in greater detail in the materials and methods section of this work. The custom electronics, computing hardware, and software infrastructure coalesce to transform a commodity vehicle into a connected automated vehicle at less than 3\% of the vehicle's retail price. This directly fills the technological gap to creating and deploying a large team of CAVs, capable of mixed autonomy traffic control.


\section*{Discussion}\label{sec:field}

    We have presented a new capability of automated vehicles: mixed autonomy traffic control. We introduce the tools and technologies for a scalable CAV fleet, validate them on a fleet of 100 vehicles, and use them on the first large scale deployment of mixed autonomy traffic control. The CAV fleet was deployed, reaching an average penetration rate of approximately 2.7\%-3.6\% per lane, during the week of November 14-18, 2022 on I-24 near Nashville, TN in heavy morning congestion. Our open scalable hardware and software platform enables experimental autonomy and connectivity in commercially available SAE level 1 and level 2 automated vehicles. Critically, our tools also enabled a rapid develop and deploy cycle. During the week, we tested multiple software versions on the fleet, enabling new controls to be run from one day to the next. 

    Our deployment of CAVs at scale differs from other field work in a fundamental way: scale. The scale of the robotic vehicle fleet changes the fundamental design of the physical platform, effecting everything upstream, including all software and control algorithms. Managing scale which is orders of magnitude larger than the baseline (one-off small-scale experimental automated vehicles of the past), is a challenge in itself. Scaling introduces a need for meta-automation, quality control, reproducibility, and monitoring software simply not necessary at small-scale. These features support the reliability of our fleet in adverse 'in the wild' field conditions. An issue previously handled in 1 minute becomes a 2 hour issue if approached with the same process. The constraints of scale met in this work forced costs (computational, financial, time) significantly down on a per-vehicle basis. These efficiencies can be carried forward into small scale field research as well. The financial barrier to entry in fielding experimental control is scaled down two orders of magnitude, something appealing to researchers interested in developing translational technologies that make a societal impact.


    This work opens up new frontiers in researching CAVs in the wild. There are several open questions about the potential effect of broad CAV adoption in real traffic flow conditions which can now be investigated, such as the effectiveness of varying controllers and their dependency of the road geometry, individual vehicle dynamics, and hyper-local traffic conditions, which all have non-trivial effects. This work allows us to shift from a somewhat myopic perspective of fielding automation in some vehicles, to a plenary perspective of changing the nature of traffic flow in the wild.



    \subsection*{Possible Extensions}
        Future work will focus on extending deployment from days into months or years; the short deployment window is a limitation in this work. A long-term deployment could have expansive new capabilities, and would need augmentation of our current tools and technologies. An ongoing long-term fleet deployment could become a mixed autonomy traffic control version of the Robotarium~\cite{pickem2017robotarium}, a remotely accessible multi-robot research testbed. A technical challenge in pursuing a long-term deployment is in privacy, security, and robustness. To address privacy in a long-term study, some preliminary work \cite{bunting2023edge} shows how edge-based privacy can be implemented, where sensitive data is not recorded. Increased security is needed in a long-term deployments since there is less physical access to the vehicles by researchers, and is necessary to prevent tampering on distributed devices. The rigors of long-term deployments would demand greater reliability and robustness from our software systems. Consider when vehicles have limited networking connections: the current software assumes a regular wireless connection for an automated recording pipeline; without rectifying this assumption major failures could occur.  Longitudinal studies in naturalistic driving vary in fleet size from around 10~\cite{varotto2022adaptive}, to 100~\cite{dingus2006100}, and then 1000s~\cite{dingus2015naturalistic,blatt2015naturalistic,benmimoun2013eurofot,takeda2011international} of deployed vehicles. They are not pointed at a specific research application or question \textit{per se}. They aim to provide a valuable resource to the research community: volumes of naturalistic driving data from a broad scope of drivers, vehicles, and locations. Outside of the research community, vehicle OEMs collect data from their fleets of vehicles which are in the millions; this data is kept private for competitive advantage and regulatory reasons. A limitation of the large research data sets is that they cannot be used to evaluate new vehicle technologies that emerge. A long-term ongoing deployment of our tools opens up the time and capability to probe the numerous open questions on the effect of new and evolving CAV technologies on transportation systems.

    Additional deployments using the tools in this work could introduce control algorithms to focus on objectives beyond traffic smoothing that are relevant to traffic engineers and society writ-large, including safety and throughput. Our technology backbone, created to field CAVs at scale, significantly lowers the barriers to fielding novel vehicle technologies in the wild in general. Already, distributed experimental sensing in the wild has been featured and adopted across domains: on roadways~\cite{eriksson2008pothole,hull2006cartel,herrera2010evaluation,work2009lagrangian}, in estuarial and riverine settings~\cite{tinka2012floating}, pastures~\cite{huircan2010zigbee}, and in production-scale fermentation processes~\cite{bisgaard2020flow}; distributed sensing combined with experimental control, shown here, is not yet pervasive. There is a history of experimental automated vehicle control at a small-scale~\cite{thrun2006stanley,jochem1993maniac, bhadani2018cat,chang1991experimentation,al2010experimental,jin2018experimental,kato2002vehicle,ma2019eco}. Instead of of pushing the ceiling of automation in vehicles higher, we look to test ideas involving distributed vehicle sensing with experimental control for improving safety and efficiency at a societal scale in the wild

    Using this work as a basis, there are many research opportunities for fielding technologies in a full-scale mixed autonomy setting. For example, investigating time-sensitive networking (TSN) between vehicles, where a guarantee on the delivery of information can be made. This guarantee could be used to augment safety assurances in car-following~\cite{gunter2022experimental}, and platooning~\cite{chang1991experimentation,al2010experimental} settings. Versions of safe message passing have been derived, simulated, and tested at small-scale~\cite{yin2004performance,xu2004vehicle,bai2006reliability}, but still need be fielded in a mixed autonomy full scale real world setting to address the sim-to-real gap.

\section*{Materials and Methods}


\begin{figure}
    \centering
    \includegraphics[width=\linewidth]{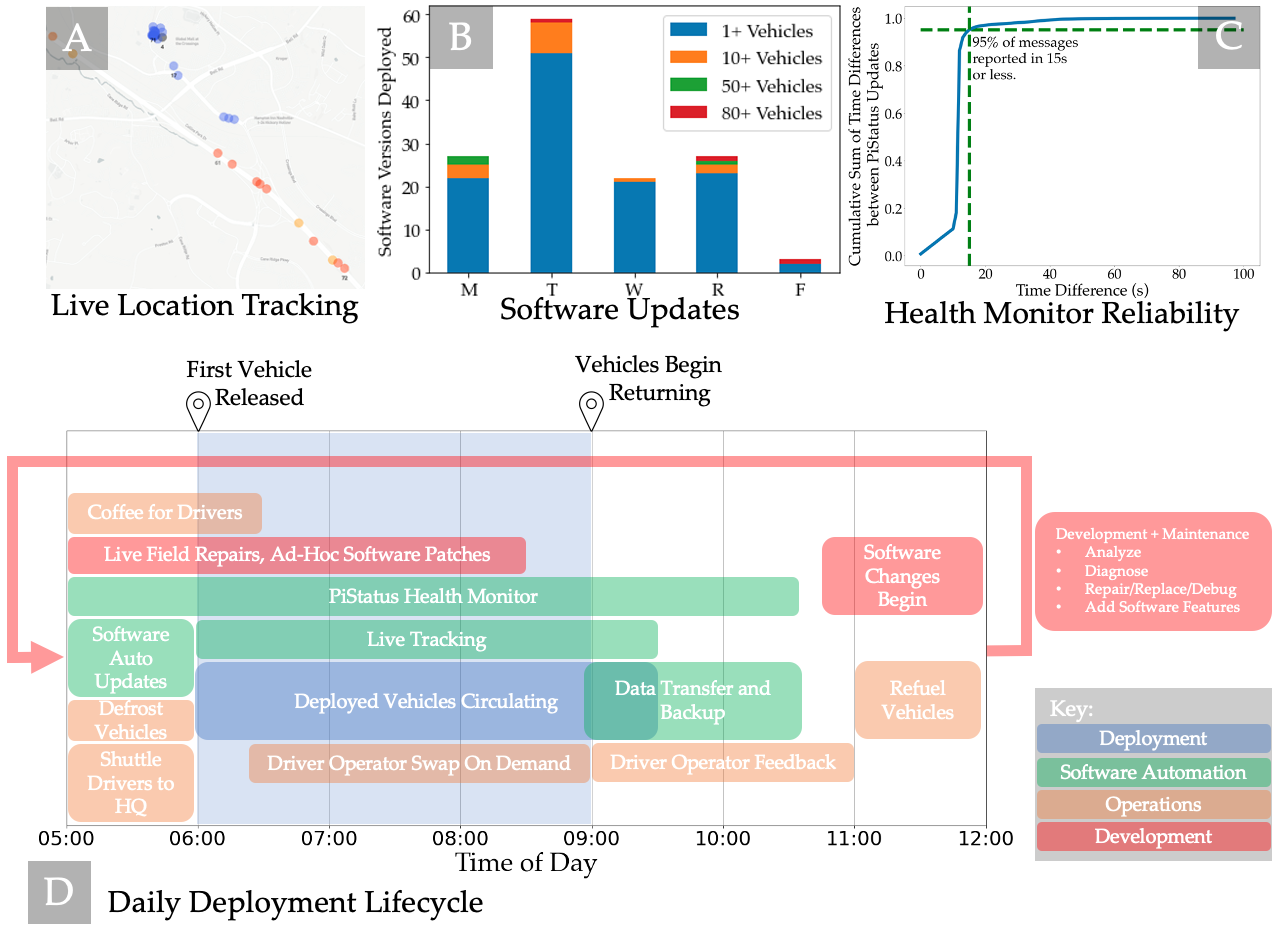}
    \caption{\textbf{Managing Scale in a 100 CAV Deployment.} Software features key to managing scale (\textbf{A-C}) are shown. Playback of (A) is featured in Movie 1. Many software versions (B) were deployed, with the most distributed ones being deployed in 10+, 50+, and 80+ vehicles. The daily lifecycle of a deployment is overviewed (\textbf{D}). Outside of the circulating vehicles, there is a collection of software automation, operations tasks, and iterative development which critically support the field deployment.}
    \label{fig:arch}
\end{figure}
\subsection*{System Architecture}\label{sec:arch}

This section provides an overview of the organization, design, and implementation of the individual vehicle platform. The system architecture is split into three categories: (i) vehicle layer, (ii) the `bridge' layers (physical and software), and (iii) server layer. The \textit{Vehicle Layer} includes the stock Advanced Driver Assistant System (ADAS) and original equipment manufacturer (OEM) sensor networks and electronic control. The vehicle layer meshes into the custom \textit{Bridge Layers}. The Bridge involves several components of a \textit{Hardware Bridge} and \textit{Software Bridge} that extend the commodity vehicle into a connected automated vehicle (CAV) at scale; the core function here is creating space for a fungible experimental controller to be plugged in. Finally the \textit{Server Layer} is where individual vehicle information is aggregated and used in centralized processes, or passed directly to other vehicles; the Server Layer also serves an important role in managing scale. 


\subsubsection*{Vehicle Layer}\label{sec:vehicle_layer}
We needed to adapt a commodity vehicle in order to have a scalable CAV fleet. To push the costs of scale down, we aimed to maximally leverage existing sensors and actuation mechanisms to minimize the extent of our platform's footprint. In contrast, an automated driving installation of the past can cost multiples of the cost of the vehicle in sensors and compute~\cite{bhadani2018cat, thrun2006stanley}. Recent AV field work in eco-driving\cite{ma2019eco}, using a similar CAN injection approach to control, uses embedded computing equipment that retails on the order of tens of thousands in USD (unacceptably expensive for our constraints). A stock vehicle needs several added abilities in order to become part of our CAV fleet. The most important are an interface with vehicle sensor data, and interface with vehicle control system, and scientific quality data collection.

A given vehicle's underlying technological systems can vary widely. The networking on the vehicle layer may use CAN, CAN FD, Flexray, Automotive ethernet, or various other communication protocols depending on the implementation of the specific make, model, model year, and trim. Communication may vary based on subsystem in the vehicle. Regardless of the implementation details, it is critical to have real-time high frequency sensor information from the vehicle in order for it to be used as a live mobile sensor, for experimental control actuation, and for scientific data value. The vehicle layer is endowed with these enhanced features through the hardware and software bridge.


\subsubsection*{Hardware Bridge Layer}\label{sec:hw_bridge}

    The Hardware Bridge creates interfaces between the three different software-driven entities (in-vehicle network, Software Bridge, Server Layer) of the system architecture. The hardware pieces marry natively independent systems to form our cohesive densely interconnected platform. Hardware interfaces are needed for reading on-board sensor data, for providing inputs to control algorithms, for sending control commands, for scientific data management, and system health monitoring. The most essential and cumbersome task for the Hardware Bridge is to provide a two-way physical translation between the electrical signals from vehicle sensors and digital signals used in the Software Bridge; this is achieved with our custom PCB and wire harnesses which in combination provide the two-way physical translation. The implemented Hardware Bridge consists of a custom printed circuit board (PCB) informally referred to as the MattHAT, a custom wiring harness cable for the vehicle CAN interface, a battery-powered uninterrupted power supply (UPS), and a 4G and wifi compatible radio for networking interfaces. The former two components are discussed next.

    \paragraph*{MattHAT}\label{sec:MattHAT}
    The MattHAT has three keystone functions within the system architecture. It translates physical CAN electrical signals with transceivers into digital signals to pass off to libpanda \cite{bunting2021libpanda} to handle; it uses an onboard digital potentiometer to provide an interface with the buttons on the Nissan Rogue cruise control system; and a relay switch to cut the physical CAN connection and inject commands into the CAN bus when needed. The device is compatible with CAN and CAN-FD, and OBD-II protocols. These features combined allowed for the MattHAT to be compatible with all vehicles in our heterogeneous fleet, and prospectively with many more vehicle platforms. The MattHAT is based on a 144-pin 32-bit STMH723ZGTb microcontroller, chosen due to its three on-board CAN FD hardware peripherals. While the MattHAT plugs into the Raspberry Pi's General Purpose Input Output (GPIO) header, USB is used as the primary connection for CAN communication due the support of CAN over USB in libpanda. The MattHAT features a few jumper selections for power selection and for initiating the firmware boot of the microcontroller. The MattHAT's digital potentiometer is an entirely separate circuit designed into the PCB.  This circuit connects the Raspberry Pi GPIO header over the Serial Peripheral Interface (SPI) supported by libpanda. The potentiometer connects to the vehicle's wire harness using a screw terminal and separate wire.  A GPIO-controlled small relay is used to arm the digital potentiometer to prevent accidental voltage inputs during startup.


    \paragraph*{Custom CAN Cable}\label{sec:can-cable}
    To gain access to vehicles' onboard sensors and pass the physical signals to/from the MattHAT, you need an access point. The CAN cables were required to carry out the encoded system state as reported on CAN (e.g. velocity, steering wheel position) and send control commands into the vehicle. Local in-vehicle sensor networks carry all information for vehicle use including the radar and sonar sensors, odometer, and even window motor control. It is the closest piece of the Hardware Bridge to the stock vehicle in the two-way pipeline. Incoming information travels from vehicle sensors to control algorithms, and then outgoing from those algorithms through the CAN cable to the vehicle for commanding control.

    Unlike the higher order layers of the system architecture, this piece of the system architecture was informed by vehicle layer detail -- connector pieces, pin locations, pin type, crimping tools, wiring insulation, twisted pair combinations, and wire length. There are several points of access on a given vehicle, with several different protocols, a myriad of partially overlapping sensor and system-state data, which vary widely from vehicle to vehicle. Our implementation ultimately relied on developing new approaches to parameter estimation in cyber-physical systems~\cite{nice2023parameter}. This made it possible to discern the critical sensor data from the noise and impertinent information. With no off-the-shelf solution, it took several hundred hours to finalize the design of the cable, and several hundred more hours to manufacture the next 100 cables.




\subsubsection*{Software Bridge Layer}\label{sec:sw_bridge}
    This layer of the system architecture sits between the Server Layer and the Hardware Bridge.
    The libpanda package~\cite{bunting2021libpanda} and the CAN to ROS tool~\cite{nice2023middleware}, which bridges in-vehicle (nominally CAN) and ROS networks, serve as the pipeline between the Hardware Bridge and the rest of the Software Bridge. The cumbersome heterogeneity of different vehicles are abstracted away by the self-configuration and code generation features of CAN to ROS to shield the upper software layers from the lower level complexity and heterogeneity in the hardware. Our Linux systemd services contribute in several different categories, acting as managers and organizers. They operate the data pipeline between the embedded devices and the Server Layer, and also manage and organize the embedded software and hardware systems. The ROS Layer of the Software Bridge is a message passing network layer to process, record, and distribute system information, including housing fungible vehicle control algorithms.

    \paragraph*{Libpanda}
    Our software package\cite{bunting2021libpanda} was used to create vehicle interfaces. It governed the use of the custom PCB for the data interfaces between the vehicle layer and the Software Bridge, and it managed the control interface with state-based models to actuate based on requests from the controller node in the ROS layer of the Software Bridge.

    Knowing that driver operators may have unexpected behaviors, and with safety as the first priority, libpanda leaves the stock safety systems intact and emulates the stock cruise control system to maximize safety and comfort despite being an experiment control system. We created supervisory control models in libpanda, translating control requests from the ROS layer into vehicle actuation, accounting for inevitable issues arising from control design integration and adverse field conditions. A control heartbeat guaranteed either fresh experimental control liveness, or stock OEM control. A whitelist from the Server Layer required central permissions in order to execute experimental control on a given vehicle. These guards held in place by libpanda work to maximize safety and performance in the field.

    \paragraph*{CAN to ROS}\label{sec:can_to_ros}
    This software package~\cite{nice2023middleware} is a two-way interface between the in-vehicle network and the ROS network. In tandem with libpanda, it creates an interface from ROS to the vehicle directly. Its model-based code generation allows for automated self-configuration at runtime enabling real-time sensing in ROS on heterogeneous vehicle systems.

    \paragraph*{Systemd services}
    Several Linux Systemd services were created to manage the embedded software systems. These software services were the mechanism by which we were able to effectuate automatic software updates, automated data management (recording and transfer), live tracking and system status monitoring, and communications with the server layer for control planning.

\paragraph*{ROS Layer}\label{sec:ros_layer}
    As a result of the CAN to ROS package, the ROS framework layer provides an approachable software space for non-domain experts to run experimental control algorithms. Vehicle interfaces for control and data are abstracted away, and what remains is a developer-friendly environment. We tutored control designers to integrate their control software in our preferred method: use a provided Simulink model of the Software Bridge, edit the controller node, and generate the formally verified ROS code. Alternatively control designers could write their controller manually in a ROS node. Either way, they were able to test their implementation with software-in-the-loop on-demand. We created a remotely accessible set of embedded hardware testing kits, each consisting of an embedded computer running the software bridge, substituting live data from the hardware bridge with pre-recorded sensor data played back through ROS. Controller designers were given remote SSH access to these kits to perform test runs with their new controllers, and compare the controller output on the pre-recorded data with their expected output.

\subsubsection*{Server Layer}\label{sec:server}
    Individual vehicles interface with the centralized server layer via the embedded compute. Each vehicle sends up local information and receives non-local information. Sometimes, local information sent up is for system health and monitoring such as the live tracking and health monitoring `PiStatus' applications, which are critical for scalability. In applications such as latent networking and control heartbeat an exchange is made between the server and individual vehicle to inform live longitudinal control of the vehicle on the open road.


\paragraph*{Latent Networking}
    Latent networking created the opportunity to build control whitelisting, a control heartbeat, and live connectivity between vehicles. Control whitelisting created a centralized power to enable or disable a vehicle's experimental control live. In case of emergency, this feature assures some operational control from the field HQ. The control heartbeat ensures liveness of connectivity from the individual vehicle for experimental control to be enabled, otherwise stock systems would be in control. This way, if for any reason a vehicle does not have live connectivity the local safety is supported. Live connectivity between vehicles was leveraged to input empirical measurements from CAN data from some vehicles into the control algorithms decisions for other vehicles. By sharing information through a central portal, traffic conditions in areas relevant to any given `ego' vehicle were available agnostic to proximity or fixed infrastructure. Experimental control implementations were empowered to decide which sets of information are or are not pertinent.   

\paragraph*{Control Heartbeat}

    Libpanda was redesigned for a heterogeneous fleet with vehicle interface abstraction in mind, extending the previous functionality of message rate checking needed for Toyota control.  The control heartbeat served as a safety measure to disconnect control messages if no service is regularly sending commands, e.g. when a control command node process crashes on the embedded compute.  This was initially built out for control in Toyota vehicles, because failing to send commands at a high enough rate will cause manufacturer modules to error for safety reasons. Libpanda had therefore been designed to check the command rate and disable control both gracefully and safely. This functionality was extended to work across all supported vehicles, which have varying liveness constraints before triggering errors in manufacturer modules. The heartbeat is also a guard against the liveness limitations of experimental algorithms. Libpanda connects the control heartbeat into ROS, so it functions as a time-dependent request. By exposing this information to the ROS layer, any higher level controllers or services can request controls to be enabled. If these higher-level functions stall or crash then libpanda will relinquish spoofing, resulting in the vehicles to operating like normal stock systems.


\subsection*{Managing Scale}\label{sec:scale}

At its core, this work deals with combatting issues that arose from efforts to scale. Societal scale problems need large scale studies to investigate their solutions, but implementation at unprecedented scale in this case requires new tools. We built tools to track live deployment of vehicle controllers, enable large-scale deployment of the electronics hardware bridge, monitor CAV system health in real-time, manage distributed software versioning, and manage data collection and transfer. These tools are necessary when fielding a large CAV fleet as a research instrument. Where automation could not be of use, we leveraged process, structure, and organization to manage complexity and performance.

The CAV fleet tools and technologies were developed with maximal modularity; this promotes co-development and adaptability. Co-development was very useful in development phases, to have many smaller components be pushed along by different team members and to maximize `uptime' when waiting for components with supply-chain constraints. Agility and flexibility are critical in large-scale field tests, as the tools and technologies run into unpredictable and chaotic conditions, and adaptability makes or breaks the ultimate functionality. We were faced with a handful of technology failures which we did not encounter in development and testing, and were saved by the adaptability built into our systems.


\paragraph*{Live Tracking}\label{sec:live-viewer}

A critical component of understanding the status of the 100 CAV team was a live tracking map, showing each vehicle and its status in real-time on a road map. We used this to understand and monitor the distribution of vehicle team within the traffic flow, the liveness of vehicle connectivity spatially throughout the testbed, which vehicles were running experimental control, their travel direction, and their velocity.
This tool is made to be leveraged by users unfamiliar with the construction of the CAV scaling platform, and lower level implementations. Control designers could use this tool to rapidly identify potential problems with novel controllers, and the impact of traffic congestion on expected test vehicle performance. It also enabled personnel teams to see at a glance how many drivers were out on the road, their location to anticipate a driver swap, and which drivers may need help. The live tracker is implemented with two pieces of software: a live-tracker ROS node running in the vehicle ROS layer, and a light-weight mapbox-based webapp. The ROS node communicates at 1 Hz with the server the following: (1) VIN (to identify the car and driver), (2) GPS coordinates and time, (3) Ego car velocity, acceleration, and control system status, (4) Whether the vehicle is moving westbound on the I-24 highway route or not. This data, when it hits the server, is promptly stored into a mySQL database for rapid retrieval. The Mapbox-based webapp contacts the server via a REST API call that returns the latest information from the database, and updates this every 15 seconds.

\paragraph*{Hardware Bridge at Scale}


Hardware scale introduces problems in the field. A task which takes 1 minute for a single vehicle will take about 2 hours for 100 vehicles. A task which takes 1 minute for a domain expert will take much longer and be performed with less accuracy by a non-domain staff member. To create a fleet of 100 CAVs, we needed staff outside the domain to do critical work in order for the system to be operational. We worked to make each task lean, easy, and intuitive. These tasks could then be monitored at scale by domain experts to focus attention on which vehicles need more work, and which are proper nominal state. 


Here we provide one detailed example to elucidate the broader thrust: We created a device to test CAN Cable quality. It consisted of the cut ends of a fully assembled CAN cable soldered to a microcontroller to systematically check the correctness of electrical connections on the handmade cables. A failure in a CAN cable has especially high cost because it requires disassembly of a vehicle to replace (see vehicle installations in Movie 1), and the production of each custom cable takes hours. When the quality control device was plugged into a computer, a display showed the correctness of the plugged in cable in a color coded wiring diagram. A spare `Gold Standard' cable was made for reference to assure the cable tester was correct. This automated quality control saved countless errors in cable manufacture, and resulted in zero failures from the CAN cables in the field.


The preparation for hardware at scale affected several other processes for field testing, both through logistics and automation. Before the fleet of 100 stock vehicles was available, we sourced several tool sets for disassembly and reassembly, documented and tested our processes for dissemination to non-domain staff, and to close our field work we were able to disassemble and uninstall our system within 2 days to return vehicles in factory-state to the manufacturer. Flashing the firmware on a MattHAT PCB was semi-automated so it could be done for all devices in under 2 hours; another example of how just making a 10 minute process 2 minutes saves days of work at scale. Since battery UPS could be overdrawn in field conditions (and cause system malfunctions if too low) the PiStatus monitor added live reporting on the battery voltage \textit{in situ}. This addition was seamless, enabled through automated software updates in the field. We could monitor the battery UPS voltage for each vehicle and by color-coded status proactively swap in batteries from a bank of fully charged ones.


\subsubsection*{PiStatus}\label{sec:pistatus}

In order to successfully process hardware installation, software installation and perpetual updates, system performance, and vehicle-by-vehicle status in the field, a live monitoring system is a necessity. This monitor also led to vehicle issue diagnoses with cables, software bugs, and more in the field. This software service, PiStatus, including its server-side database counterpart, allowed for real time monitoring of the health granular enough for each installed component, and broad enough for the entire CAV team functionality.

In implementation, PiStatus is a novel systemd service for self-reporting vehicles' status. Bash scripting and other interconnected systemd services provided the local information from each vehicle to be sent to the server-side database 5 times per minute; from this database, several php webpages were created to display and organize this information.  Small details, such as color coding the information displayed on the site, made it possible for a holistic understanding of the distributed system in the blink of an eye. 

\subsubsection*{Software Updates} \label{sec:updates}

Automatically deploying software upgrades is an essential function of the fleet as a research instrument. Foremost, this is the critical difference between deploying an algorithm vs. any algorithm for mixed autonomy traffic control. Second, the inevitability of software failures must be planned for, since a complete fidelity simulator for the physical deployment of CAVs does not exist. A control designer without domain expertise in robotics, automation, or vehicle systems, must be able to create a new or updated control system and deploy it for testing immediately, otherwise there is no realistic chance for the fleet to be properly calibrated and controlled in the critical time window of field testing.
To collapse the development/deployment feedback cycle we needed a mechanism to update controllers systematically and on-demand. Our design choice was for the embedded device to look for an update when powered on, and install updates immediately if there is an update to be made. Each update includes a prescribed set of software version hashes to all repositories' versions to be used. There were multiple of these sets of hashes, assigned by VIN, to allow for running multiple versions of control simultaneously in the fleet. By creating the ability to vary software version on a vehicle-by-vehicle basis, we created the opportunity for partitions of the fleet to be tested A/B and for unstable changes to be hardened. Software updates were managed through git, with some critical changes happening down to the minute with coordinated radio calls, and others on a scheduled basis.
As a practical matter, software updates were managed by closing access to the libpanda repository in the organization to all but the core team. A mismatch between a new commit on the master branch and the local version triggered updates (pull and rebuild). To ensure beyond a doubt that systems were up to the minute with bug fixes, we often had to `boot' all vehicles. Since we created a software version monitor in PiStatus, we could confirm a vehicle's software version before letting them drive. The update attempt would occur without this redundant check, but monitoring gave assurance that no runtime issues prevented the software installations. Updates occurred ad-hoc during testing, at night to prepare for the next day, the morning of, and in the afternoons to test new software in small-to-medium scale tests.

\subsubsection*{Data Management}\label{sec:data_management}

    \paragraph*{Scientific Data Quality}
        The vehicles are collecting high fidelity information about their state at 20+ Hz. This richness is critical for applications which need to understand dynamics such as controls, traffic engineering and micro-simulation, and artificial intelligence research~\cite{lichtle2022deploying,gunter2022experimental}. The archival data collection enables analyses on the effectiveness of each different controller fielded, and several other research questions of interest. These include: studying the effects of the vehicle operator as a human-in-the-loop; analyzing the effectiveness of software upgrades; or collected data used as ground truth to compare to camera tracked trajectories. There already exists an open sourced tool for analysis of CAN data\cite{bhadani2022strym} which makes CAN analysis approachable. This first deployment of mixed autonomy traffic control has generated 22752 miles of driving data over a combined 1022 hours, a novel and valuable data archive.

    \paragraph*{Automated Data Transfer}
        Data collection is part of the automated startup of the operating system on boot. We collect CAN data to capture the in-vehicle network in high fidelity, GPS data to capture location and velocity redundantly, rosbag files to capture the ROS network traffic, system logs, and launch environment JSON files for post-hoc analyses. A total of 13.7 Terabytes of data were collected over 5 days of driving. Over 1.7 Terabytes of which was vehicle CAN data, GPS, and ROS bagfiles. The remaining 12 Terabytes are dashcam video data. We used one private/public key pair embedded in a locally distributed OS copy to allow each RPi device access to a user account for transferring data to our local server. We set up a hardened wireless access point in the secure parking area for the fleet, spliced from optical fibers of our operational hub. We also created a tool to monitor the data transfer live at scale from the server-side. On the hour, these data were synced to the Cyverse\cite{swetnam2023cyverse} data store in the cloud. Having the data recorded and uploaded automatically saved countless hours of work to assure start/stop of records and manually transfer the data. Since the data are so critical for the controller developers to examine the results of their algorithms' deployment, it was useful to know that the data were in fact in hand, properly recorded and backed up automatically. Furthermore, because of the dynamic nature of our system, data were examined in the morning of the field deployments and new controllers were being tested hours later for the next days 100 CAV fleet deployment. Fresh data informed critical field decision-making. This capability was enabled in part through automated data transfer.

\subsubsection*{Process Vs. Automation}

During field testing, there were greater than 50 non-domain field staff members but at most 4 personnel had enough domain knowledge to fix the entire vehicle platform. Often only 1 or 2 researchers had the required `full stack' knowledge. Since there were too many tasks to do manually, automation proved essential to the functionality of the CAV fleet. An equilibrium state developed between manual work and automation development costs; this was applied to even the smallest tasks. For example, physical vehicle key management for the fleet was managed by organizational process using a simple paper system. However, battery charge state was managed automatically through PiStatus because checking voltage one-by-one manually is much too costly. As much as possible, one should avoid needing to explain domain-specific knowledge to non-domain experts in the field. It is better to commit to a tried process, or automate it with humans out-of-the-loop. Since there is no road map suggesting where process vs. automation is more valuable for managing a fleet of 100 CAVs, because it has never been done before, we were empowered to invent along the way.

\subsubsection*{Repairs and Technology Failures}
         It was a surprise how vital our PiStatus tool, the live system monitor, was to the success of the field deployment in dealing with repairs and technology failures. We would have not been able to successfully install, maintain, and update all of the cables and embedded devices on all of the vehicles without a live and continuous monitoring tool detailing the state of the electronics and software. Some third-party hardware failed or failed to perform at times in the field. We observed inconsistent errors from SD cards, resulting in empty binaries or non-bootable operating systems, and some third-party electronics with subpar quality simply failed. Another hurdle was loose or unplugged cable connections, as there were several connection points per vehicle so several hundred opportunities for a loose or incomplete connection. Some missed cable connections were a result of the rapid disassembly/reassembly of the vehicle, causing temporary OEM system failures (e.g. parking brake non-functional). Each cable connection is a single-point failure for a vehicle's CAV functionality, so 100\% compliance was necessary.

        Some failures were triggered only once the large-scale field testing environment was introduced. For example, the UPS boards worked without issue in testing for auto-on functionality when external power supplied. We did not anticipate the high volume that the vehicles were going to be power cycling (from defrosting unseasonable ice, updating software at the last minute, field staff testing system functions). This revealed a quirk in the UPS boards where a capacitor in the `Auto-On' circuit had not yet discharged when the vehicle was power cycled, rendering a conditional failure in auto-on functionality. 

\bibliography{references}
\bibliographystyle{Science}

\section*{Acknowledgments}
The authors would like to thank Nissan, Toyota, General Motors, and the Tennessee Department of Transportation for their partnership on this work. This material is based upon work supported by the U.S. Department of Energy’s Office of Energy Efficiency and Renewable Energy (EERE) award number CID DE-EE0008872, and the National Science Foundation awards 1837652, 1837481, and 2135579. The views expressed herein do not necessarily represent the views of the U.S. Department of Energy or the United States Government. This work is supported by the Dwight D. Eisenhower Fellowship program under Grant No. 693JJ32345023 (Nice).

\end{document}